\pdfoutput=1
\documentclass[11pt,a4paper]{article}
\usepackage[hyperref]{eacl2020}
\usepackage{times}
\usepackage{latexsym}

\usepackage{hyperref}
\usepackage{xcolor}
\usepackage{subcaption}
\usepackage{comment}
\usepackage{booktabs}
\usepackage{graphicx,multirow}
\usepackage{amsmath}
\usepackage{lipsum}
\usepackage{enumitem}

\usepackage{microtype}

\aclfinalcopy 

\title{
Building Representative Corpora from Illiterate Communities:
A Review of Challenges and Mitigation Strategies for Developing Countries
\\
}

\author{
  Stephanie Hirmer$^{1}$,
  Alycia Leonard$^{1}$,
  Josephine Tumwesige$^{2}$,
  Costanza Conforti$^{2,3}$\\
  $^1$Energy and Power Group, University of Oxford\\
  $^2$Rural Senses Ltd.\\
  $^3$Language Technology Lab, University of Cambridge\\
  {\tt stephanie.hirmer@eng.ox.ac.uk}
  }
\date{}

\begin{document}
\maketitle
\begin{abstract}
%
%
Most well-established data collection methods currently adopted in NLP depend on the assumption of speaker literacy.
%
Consequently, the collected corpora largely fail to represent swathes of the global population, which tend to be some of the most vulnerable and marginalised people in society, and often live in rural developing areas.
%
Such underrepresented groups are thus not only ignored when making modeling and system design decisions, but also prevented from benefiting from 
development outcomes achieved through data-driven NLP.
%
%
This paper aims to address the under-representation of illiterate 
communities in NLP corpora: 
we
identify potential biases and ethical issues that might  arise when collecting data from rural communities with high illiteracy rates in Low-Income Countries, and propose a set of practical mitigation strategies to help future work. 

\end{abstract}

\section{Introduction}
\begin{figure*}[t]
    \centering
    \begin{subfigure}[b]{0.3\textwidth}
        \centering
        \includegraphics[width=\textwidth]{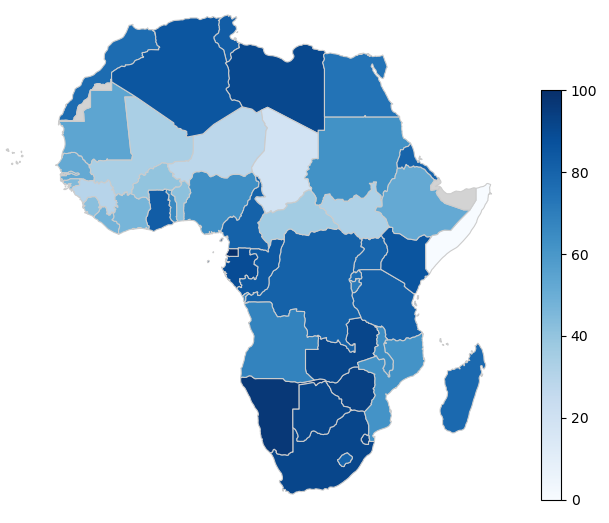}
        \caption{Adult literacy 
        (\% ages 15+, \citet{UNESCO:2018})}
        \label{fig:literacy}
    \end{subfigure}
    \hfill
    \begin{subfigure}[b]{0.3\textwidth}
        \centering
        \includegraphics[width=\textwidth]{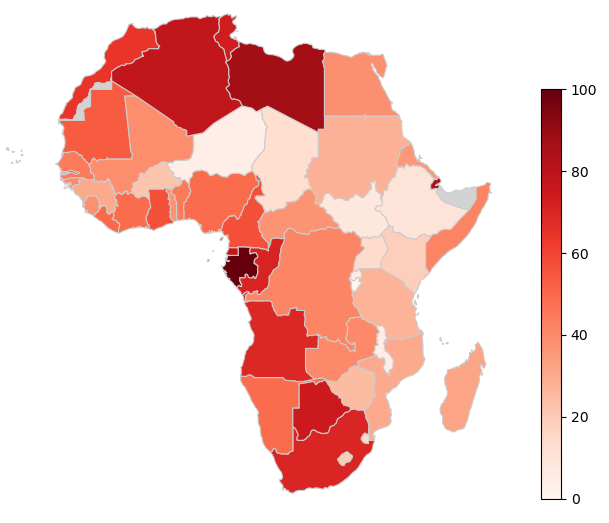}
        \caption{Urban population (\% total, \citet{UNDESA:2018})}
        \label{fig:urban}
    \end{subfigure}
    \hfill
    \begin{subfigure}[b]{0.3\textwidth}
        \centering
        \includegraphics[width=\textwidth]{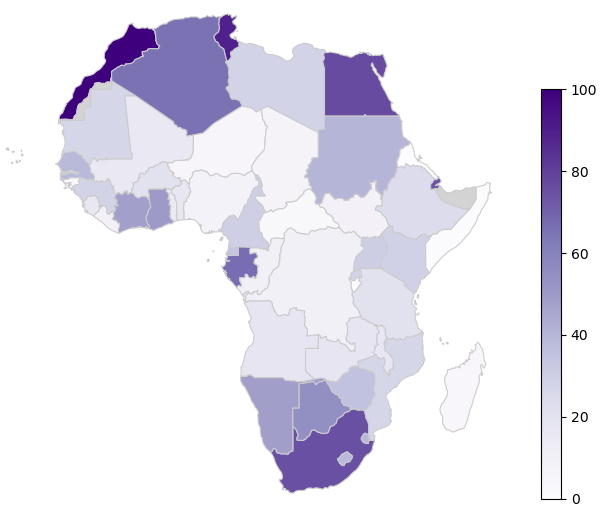}
        \caption{Internet usage (\% of total, \citet{ITU:2019})}
        \label{fig:internet}
    \end{subfigure}
    \caption{Literacy, urban population, and internet usage in African countries. 
    Note 
    that countries with more rural populations tend to have less literacy and less internet users. These countries are likely to be under-represented in corpora generated using common data collection methods that assume literacy and internet access (Grey: no data).}
    \label{fig:map_africa}
\end{figure*}

The exponentially increasing popularity of supervised Machine Learning (ML) in the past decade has made the availability of data crucial to the development of the Natural Language Processing (NLP) field.
As a result, much NLP research has focused on developing rigorous processes for collecting large corpora suitable for training ML systems. 
We observe, however, that many 
best practices 
for quality data collection make two implicit assumptions:
%
that speakers have {internet access} and
that they are {literate} (i.e. able to read and often write text effortlessly\footnote{For example, input from speakers is often taken in writing, in response to a written stimulus which must be read.}).
%
Such assumptions might be reasonable in the context of most High-Income Countries (HICs)~\cite{UNESCO:2018}. 
%
However, in 
Low-Income Countries (LICs), and especially in sub-Saharan Africa (SSA), such assumptions may not hold, particularly in rural developing areas where the bulk of the population lives~(\citet{owidliteracy}, Figure~\ref{fig:map_africa}).
%
As a consequence, common data collection techniques – designed for use in HICs – 
fail to capture data from a vast portion of the population when applied to LICs.
%
Such techniques include, for example, crowdsourcing~\cite{packham2016crowdsourcing},  
scraping social media~\cite{le2016sentiment}
or other websites~\cite{roy-etal-2020-topic},  
collecting articles from local newspapers~\cite{marivate2020investigating}, or interviewing experts from international organizations~\cite{friedman2017multi}.
While these techniques are important to easily build large corpora, they implicitly rely on the above-mentioned assumptions (i.e.~internet access and literacy), and might result in {demographic misrepresentation}~\cite{hovy-spruit-2016-social}.
In this paper, we make a first step towards addressing 
\textit{how to build representative corpora in 
LICs from illiterate speakers}. We believe that this is a currently unaddressed 
topic within NLP research.
It aligns with previous work investigating sources of bias resulting from the under-representation of specific demographic groups in NLP corpora (such as women~\cite{hovy-2015-demographic}, youth~\cite{hovy-sogaard-2015-tagging}, or ethnic minorities~\cite{groenwold-etal-2020-investigating}).
%
%
%
%
%
%
%
In this paper, we make the following contributions:
(i)~we introduce the challenges of collecting data from illiterate speakers in~$\mathsection$\ref{challenges}; 
(ii)~
we define various possible sources of biases and ethical issues which can contribute to low data quality 
we define various possible sources of biases and ethical issues which can contribute to low data quality 
we define various possible sources of biases and ethical issues which can contribute to low data quality 
$\mathsection$\ref{defs}; finally,
(iii)~drawing on years of experience in data collection in 
LICs, we outline practical countermeasures to address these issues in $\mathsection$\ref{countermeasures}. 
\section{Listening to the Illiterate: What Makes it Challenging?}\label{challenges}

In recent years, 
developing corpora that encompasses
as many human languages as possible has been recognised as important in the NLP community.
%
In this context, widely translated texts (such as the Bible~\cite{mueller2020analysis} or the Human Rights declaration~\cite{king2015practical}) are often used as a source of data.
However, these texts tend to be
quite
short and domain-specific.
Moreover, 
while the Internet constitutes a powerful data collection tool which is more representative of real language use than the previously-mentioned texts,
it excludes illiterate communities, as well as speakers which lack reliable internet access (as is often the case in rural developing settings,~Figure~\ref{fig:map_africa}).

\textcolor{black}{Given the obstacles to using these common language data collection methods in LIC contexts, the NLP community can learn from methodologies adopted in other fields.}
Researchers from fields such as \textcolor{black}{sustainable development} (SD,~\citet{gleitsmann2007analysis}), African studies~\cite{adams2014decolonizing}, and ethnology~\cite{skinner2013interview}, tend to rely heavily on qualitative data from 
oral interviews, transcribed verbatim. 
Collecting such data in rural developing areas is considerably more difficult than in developed or urban contexts. 
In addition to high illiteracy levels, researchers face challenges such as seasonal roads and low population densities. 
%
%
To our knowledge, there are very few NLP works which explicitly focus on building corpora from rural and illiterate communities:
of those works that exist, some
present clear priming effect issues~\cite{abraham-etal-2020-crowdsourcing}, while others focus on application~\cite{our_emnlp}.
A detailed description of best practices for data collection remains a notable research gap.
\section{Definitions and Challenges}\label{defs}
Guided by research in 
medicine~\cite{pannucci2010identifying}, sociology~\cite{berk1983introduction}, and psychology~\cite{gilovich2002heuristics},
NLP has experienced increasing interest in ethics and bias mitigation to minimise unintentional demographic misrepresentation and harm~\cite{hovy-spruit-2016-social}.
While there are many stages where bias may enter the NLP pipeline~\cite{shah2019predictive}, we focus on those pertinent to 
\textit{data collection} from rural illiterate communities in LICs, 
\textcolor{black}{leaving the study of biases in 
model development for future work}\footnote{Note, this paper does not focus on a particular NLP application, as once the data has been collected from illiterate communities it can be annotated for virtually any specific task.}.

\subsection{Data Collection Biases}
\noindent
Biases in data collection are inevitable \cite{Marshall1996} but can be minimised when known to the researcher \cite{Trembley1957}.
We identify various 
biases that can emerge when collecting language 
data in rural developing contexts, which fall under three broad categories: sampling, observer, and response bias.
{Sampling} determines {who} is studied, the {interviewer} (or observer) determines {what information} is sought and {how} it is interpreted, and the {interviewee} (or respondent) determines 
which information is revealed~\cite{Woodhouse1998a}. 
These categories span the entire data collection process and can affect the quality and quantity of language data obtained. 
%
%

%

%
%

%

\subsection{Sampling or selection bias}
Sampling bias occurs when observations are drawn from an unrepresentative subset of the population being studied 
~\cite{Marshall1996} and applied more widely.
In our context, this might arise \textcolor{black}{when selecting communities from which to collect language data, or specific individuals within each community.} 
When sampling communities, 
bias can be introduced 
if convenience is prioritized. \textcolor{black}{Communities which are easier to access may not produce language data representative of a larger area or group.} This can be illustrated through Uganda's refugee response, which consists of  13 settlements (including the 2nd largest in the world) hosted in 12 districts \cite{UNHCR:2020}. Data collection may be easier in 
one of the  older, established settlements; 
however, such data cannot be generalised over the entire refugee response due to different cultural backgrounds, length of stay of refugees in different areas, 
and the varied stages along the humanitarian chain – emergency, recovery or development – found therein ~\cite{Winter1983,OECD:2019}. \textcolor{black}{Prioritizing convenience in this case may result in  corpora which over-represents the cultural and economic contexts of more established, longer-term refugees}.
%
%
%
When sampling interviewees,
bias can be introduced when certain sub-sets of a community have more data collected than others \cite{Bryman2012c}. This is seen when data is
collected only from men in a community due to cultural norms \cite{Nadal2017},
or only
from wealthier people in cell-phone-based surveys~\cite{labrique2017health}. 

%
%

\subsubsection{Observer bias}
Observer bias occurs when there are systematic errors in how data is recorded, 
which may stem from observer viewpoints and predispositions \cite{gonsamo2014citizen}. We identify three key observer biases relevant to our context. 

\par
Firstly, \textbf{confirmation bias}, which 
refers to the tendency to look for information which confirms one's preconceptions or hypotheses \cite{nickerson1998confirmation}. Researchers collecting data in 
LICs may expect interviewees to express needs or hardships based on their preconceptions. As \citet{kumar1987conducting} points out, ``often they hear what they want to hear and ignore what they do not want to hear''. A team conducting a needs assessment for a rural electrification project, for instance, may expect a need for electricity, and thus consciously or subconsciously seek data which confirms this, interpret potentially unrelated data as electricity-motivated \cite{hirmer2017benefits}, or omit data which contradicts their hypothesis~\cite{peters2020function}. \textcolor{black}{Using such data to train NLP models may introduce unintentional bias towards the original expectations of the researchers instead of accurately representing 
the community.}

\par
Secondly, the interviewer's understanding and interpretation of the speaker's utterances might be influenced by their class, culture and language. 
Note that, particularly in countries without strong language standardisation policies, consistent semantic shifts can happen even between varieties spoken in neighboring regions~\cite{gordon2019language}, which
may result 
in systematic \textbf{misunderstanding}~\cite{sayer2013misunderstanding}.
For example, in the neighboring Ugandan tribes of {Toro} and Bunyoro, the same 
word \textit{omunyoro} means respectively \textit{husband} and \textit{a member of the tribe}.
Language data collected in such contexts, if not properly handled,
may contain inaccuracies which lead to NLP models that misrepresent these tribes.
Rich information communicated through gesture, expression, and tone (i.e. nonverbal data, \citet{oliver2005constraints}) may also be systematically lost during verbatim transcription, causing inadvertent inconsistencies in the corpora. 
%
%

\par
Thirdly, \textbf{interviewer bias}, which refers to the subjectivity unconsciously introduced into data gathering by the worldview of the interviewer~\cite{frey2018sage}. 
For instance, a deeply religious interviewer may 
unintentionally frame questions through religious language (e.g. \textit{it is God's will}, \textit{thank God}, etc.), or may perceive certain emotions (e.g. thankfulness) as inherently religious, and record language data including this perception. The researcher's attitude and behaviour may also influence responses ~\cite{Silverman2013}; for instance,
when interviewers take longer to deliver questions, interviewees tend to provide longer responses \cite{matarazzo1963interviewer}. \textcolor{black}{Unlike in internet-based language data collection, where all speakers are exposed to uniform, text-based interfaces, collecting data from illiterate communities necessitates the presence of an interviewer, who cannot always be the same person due to scalability constraints, introducing this inevitable variability and subsequent data bias.}

\subsubsection{Response bias}

Response bias occurs when speakers provide inaccurate or false responses to questions. \textcolor{black}{This is particularly important when working in rural settings, where the majority of data collection is currently related to SD projects. The majority of existing data is biased by the projects for which it has been collected, and any newly collected data for NLP uses is also likely to be used in decision making for SD. This inherent link of data collection to material development outcomes inevitably affects what is communicated}. There are five key response biases relevant to our context. 

\par
Firstly, \textbf{recall bias}, where speakers recall only certain events or omit details \cite{coughlin1990recall}. This is often as a result of external influences, such as the presence of a data collector who is new to the community. Recall can also be affected by the distortion or amplification of traumatic memories \cite{strange2015memory}; if data is collected around a topic a speaker may find traumatic, recall bias may be unintentionally introduced. 

\par
Secondly, \textbf{social desirability bias}, which refers to the tendency of interviewees to provide socially desirable/acceptable responses rather than honest responses, particularly in certain interview contexts
\cite{bergen2020everything}. 
In tight-knit rural communities,  
it may be difficult to deviate from traditional social norms, leading to biased data. As an illustrative example, researchers in Nepal found that interviewer gender affected the detail in responses to some sensitive questions (e.g. sex and contraception): participants provided less detail to male interviewers \cite{axinn1991influence}. Social desirability bias can produce corpora which misrepresent community social dynamics or under-represent sensitive topics. 
%

\par
Thirdly, \textbf{recency effect or serial-position}, which is the tendency of a person to recall the first and last items in a series best, and the middle items worst~\cite{Troyer2011}. 
\textcolor{black}{This can greatly impact the content of language data. For instance, in the context of data collection to guide development work,} it is important to understand current needs and values \cite{hirmer2016identifying}; however, if only the most recent needs are discussed, long-term needs may be overlooked. To illustrate, while a community which has just experienced a poor agricultural season may tend to express the importance of improving agricultural output, other needs which are less top-of-mind (i.e. healthcare, education) may be equally important despite being expressed less frequently. If data containing \textit{recency bias} is used to develop NLP models, particularly for sustainable development applications~ (such as for Automatic UPV Classification,~\citet{our_emnlp}), these may amplify current needs and under-represent long-term needs. 

\par
Fourthly, \textbf{acquiescence bias}, also known as ``yea'' saying \cite{laajaj2017measuring}, which can occur in rural developing contexts when interviewees perceive that certain (possibly false) responses will 
please a data collector and bring benefits to their community. For example, 
\textcolor{black}{if data collection is being undertaken by a group} with a stated desire to build a school may be more likely to hear about how much education is valued.
%

\par
Finally, \textbf{priming effect}, or the ability of a presented stimulus to influence one's response to a subsequent stimulus~\cite{lavrakas2008encyclopedia}.
Priming is problematic in 
data collection to inform SD projects;
it can be difficult to collect data on the relative importance of simultaneous (or conflicting) needs if the community is primed to focus on one \cite{veltkamp2011motivating}.
An example is shown in Figure \ref{fig:visualprompt}; 
respondents may be drawn to speak more about the most dominant prompts presented in the chart.
This is typical of a broader failure in SD to uncover beneficiary priorities without introducing project bias~\cite{watkins2012priming}. Needs assessments, like the one referenced above linked to a rural electrification project, tend to focus explicitly on project-related needs 
instead of more broadly identifying what may be most important to communities \cite{Masangwi2015,USAID2006}. As 
speakers will usually know why 
data is being collected in such cases, 
they may be biased towards stating the project aim as a need, \textcolor{black}{thereby skewing the corpora to over-represent this aim}.
%
%
\subsection{Ethical Considerations}
\noindent
Certain ethical codes of conduct must be followed when 
collecting data from illiterate speakers in rural communities in LICs 
\cite{musoke2020ethical}.
Unethical data collection may harm communities, treat them without dignity, disrupt their lives, damage intra-community or external relationships, and disregard community norms 
\cite{DFIDethics}. 
This is particularly critical in rural developing regions, as these areas are home to some of the world's poorest and most vulnerable to exploitation~\cite{christiaensen2005towards,Cenival2008}. 
Unethical data collection can replicate extractive colonial relationships whereby data is extracted from communities with no mutual benefit or ownership \cite{dunbar2006ethics}. It can lead to a lack of trust between data collector and interviewees
and unwillingness to participate in future research 
\cite{clark2008we}. These phenomena can bias data
or 
reduce data availability.
Ethical data collection practices 
in rural developing regions with high illiteracy include: obtaining consent \cite{mcadam2004ethics}, accounting for cultural differences \cite{Silverman2013}, ensuring anonymity and confidentiality \cite{Bryman2012c}, respecting existing community or leadership structures \cite{harding2012conducting}, and making the community the owner of the data. While the latter is not often currently 
practiced, it is an important consideration for community empowerment, with indigenous data sovereignty efforts~\cite{rainie2019indigenous} 
already setting precedent.

\section{Countermeasures} \label{countermeasures}
%
Drawing on existing literature and years of field experience collecting spoken data in LICs, below we outline a number of practical data collection strategies 
to 
minimise previously-outlined
challenges ($\mathsection$\ref{defs}), \textcolor{black}{enabling the collection of high-quality, minimally-biased data from illiterate speakers in LICs suitable for use in NLP models.} 
%
%
While these measures have primarily been applied in SSA, we have also successfully tested them in projects focusing on refugees in the Middle East and rural communities in South Asia.


\begin{table*}[t]
    \centering\small
    \begin{tabular}{p{2pt}p{225pt}p{200pt}}
    \toprule
    & Bias \& Definition & Key countermeasures \\
    \midrule
    \multirow{2}{*}{\rotatebox{90}{Sampling\textcolor{white}{-}}} 
    & \textbf{Community}: An unrepresentative sample set is generalised over the entire case being studied.  
    & $\bullet$ Select representative communities
    \& only apply data within same scope (i.e. consult data statements)
    \\
    & \textbf{Participant}: Certain sub-sets of a community have more data collected from them than others.
    & $\bullet$ Select representative participants, only apply data within same scope \& avoid tempting rewards
    \\
    \midrule
    \multirow{3}{*}{\rotatebox{90}{Observer\textcolor{white}{----}}}
    & \textbf{Confirmation}: Looking for information that confirms one’s preconceptions or hypotheses about a topic/research/sector.
    & $\bullet$ Employ interviewers that are impartial to the topic/research/sector investigated.
    \\
    & \textbf{Misunderstanding}: Data is incorrectly transcribed or categorized as a result of class, cultural, or linguistic differences.
    & $\bullet$ Employ local people \& minimise 
    \# of people involved for both data collection \& transcription.
    \\
    & \textbf{Interviewer}: Unconscious subjectivity introduced into data gathering by interviewers' worldview.
    & $\bullet$ Undertake training to minimise influence exerted from questions, technology, \& attitudes.
    \\
    \midrule
    \multirow{5}{*}{\rotatebox[origin=c]{90}{Response\textcolor{white}{-----------}}}
    & \textbf{Recall}: Tendency of speakers recall only certain events or omit details
    & $\bullet$ Collect support data (e.g. from socio-economic data or local stakeholders) to compare with interviews.
    \\
    & \textbf{Social-desirability}:Tendency  of  participants to provide socially desirable/acceptable responses rather than to respond honestly.
    & $\bullet$ Select interviewers \& design interview processes to account for known norms which might skew responses 
    \\
    & \textbf{Recency effect}: Tendency to recall first or last items in a series best, \& middle items worst.
    & $\bullet$ Minimise external influence on participants throughout data gathering (e.g. technologies, people, perceptions).
    \\
    & \textbf{Acquiescence}: Respondents perceive certain, perhaps false, answers may please data collectors, bringing community benefits.
    & $\bullet$ Gather non-sectoral holistic insights (e.g. from socio-economic data or local stakeholders)
    \\
    & \textbf{Priming effect}: Ability of a presented stimulus to influence one’s response to a subsequent stimulus
    & $\bullet$ Use appropriate visual prompts (graphically similar), language and technology 
    \\
    \bottomrule
    \end{tabular}
    \caption{\textcolor{black}{Sources of potential bias 
    in data collection when operating in rural and illiterate settings in developing countries,
    and key countermeasures that can help mitigating them.
    }}
    \label{tab:biases_table}
\end{table*}


\subsection{Preparation}\label{prep}
Here, we outline practical preparation steps for careful planning, which 
can minimise error and reduce fieldwork duration~\cite{Tukey1980need}. 

\par
\textbf{Local Context}.~A thorough understanding of local context is key to successful
data collection 
\cite{Hentschel1999contextuality,Bukenya2012understanding, Launiala2006importance}. Local context is broadly defined as facts, concepts, beliefs, values, and perceptions used by local people to interpret the world around them, and is shaped by their surroundings (i.e.~their worldview, \citet{vasconcellos2014knowledge}). It is important to consider local context when preparing to collect data in rural developing areas, as common data collection methods may be inappropriate 
due to contextual linguistic differences and deep-rooted social and cultural norms \cite{Walker2011social, Mafuta2016local,Nikulina2019lost, Wang2020genomes}. 
\textcolor{black}{Selecting a contextually-appropriate data collection method} is critical in mitigating \textit{social desirability bias} \textcolor{black}{in the collected data}, among other challenges. 
%
Researchers should 
review socio-economic surveys and/or consult local stakeholders who can offer valuable insights on practices and social norms. These stakeholders can also highlight current or historical matters of concern to the area, which may be unfamiliar to researchers, and reveal local, traditional, and indigenous knowledge 
which may impact the data being collected \cite{Wu2014embedd} and result in \textit{recency effect}.
It is good practice to identify local 
conflicts and segmentation within a community,
especially in a rural context, where 
the population is vulnerable and systematically unheard \cite{Dudwick2006Bank,Mallick2011social}. 


\textbf{Case sampling}.~In qualitative research, sample cases are often strategically selected based on the research question (i.e. \textit{systematic} or \textit{purposive} sampling, \citet{Bryman2012c}), and characteristics or circumstances relevant to the topic of study \cite{yach1992use}.
If data collected in such research is used beyond its original scope, 
\textit{sampling bias} may result.
So, while data collected in previous research should be re-used \textcolor{black}{to expand NLP corpora} where possible, it is important to be cognizant of the purposive sampling underlying existing data. 
A comprehensive
dataset characterisation~\cite{bender-friedman-2018-data,gebru2018datasheets} can help researchers understand whether an existing dataset is appropriate to use 
in new or different research, \textcolor{black}{such as in training new NLP models,} and can highlight the potential ethical concerns of data re-use.
\par
\textbf{Participant sampling.}~Interviewees should be selected to represent the diverse interests of a community or sampling group (e.g.~
occupation, age, gender, religion, ethnicity or male/female household heads~\cite{Bryman2012c}) to reduce \textit{sampling bias} \cite{Kitzinger1994}. To ensure representativity in collected data, sampling should be random, i.e.~every subject has equal probability to be included 
\cite{etikan2016comparison}. 
There may be certain societal subsets that are concealed from view (e.g. as a result of embarrassment from disabilities or physical differences) based on cultural norms in less inclusive societies \cite{Vesper2019}; particular care should be exercised to ensure such subsets are represented. 
%
%

\noindent\textbf{Group composition}. Participant sampling best practices vary by data collection method, with particular care being necessary in group settings. 
In traditional societies where strong power dynamics exist, attention should be paid to group composition and interaction to prevent some voices from being silenced or over-represented \cite{stewart2007focus}. For example, in Uganda, female 
interviewees may be less likely to voice opinions in the presence of male interviewees 
\cite{FIDH2012,axinn1991influence}, introducing a form of \textit{social desirability bias} \textcolor{black}{ in resulting corpora}. 
To minimise this risk of data bias, 
relations and power dynamics 
must be considered 
during 
data collection planning~\cite{hirmerPhD}. It may be necessary to exclude, for instance, close relatives, 
governmental officials, 
and village leaders from 
group discussions \textcolor{black}{where data is being collected,} 
and instead engage such stakeholders in 
separate activities to ensure that \textcolor{black}{their voices are included in the corpora without biasing the data collected from others.}
    
\textbf{Interviewer selection}.~The interviewer has a significant opportunity to introduce \textit{observer and response biases} in collected data \cite{salazar1990interviewer}. 
Interviewers familiar with local language, including community-specific dialects, should be selected wherever possible. Moreover, 
to reduce \textit{misunderstanding} and \textit{recall biases} in collected data, it is useful to have the same person who conducts the interviews also transcribe them. This minimizes the layers of linguistic interpretation affecting the final dataset and can increase accuracy through familiarity with the interview content. If the interviewer is unavailable, the transcriber must be properly trained and briefed on the interviews, and made aware of the level of detail needed during transcription \cite{parcell2017interviews}. 

\begin{figure*}[t!]
\begin{subfigure}[b]{0.36\textwidth}
 \centering
 \includegraphics[width=5.5cm]{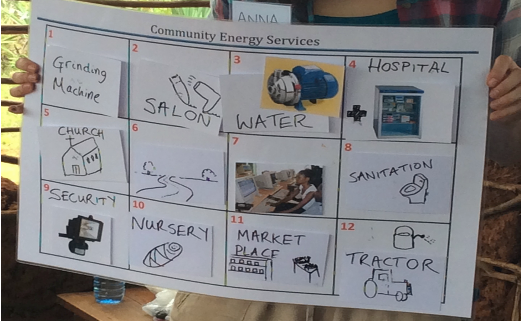}
    \caption{}
    \label{fig:visualprompt}
 \end{subfigure}
 \hfill
 \begin{subfigure}[b]{0.25\textwidth}
    \centering
    \includegraphics[width=4.2cm]{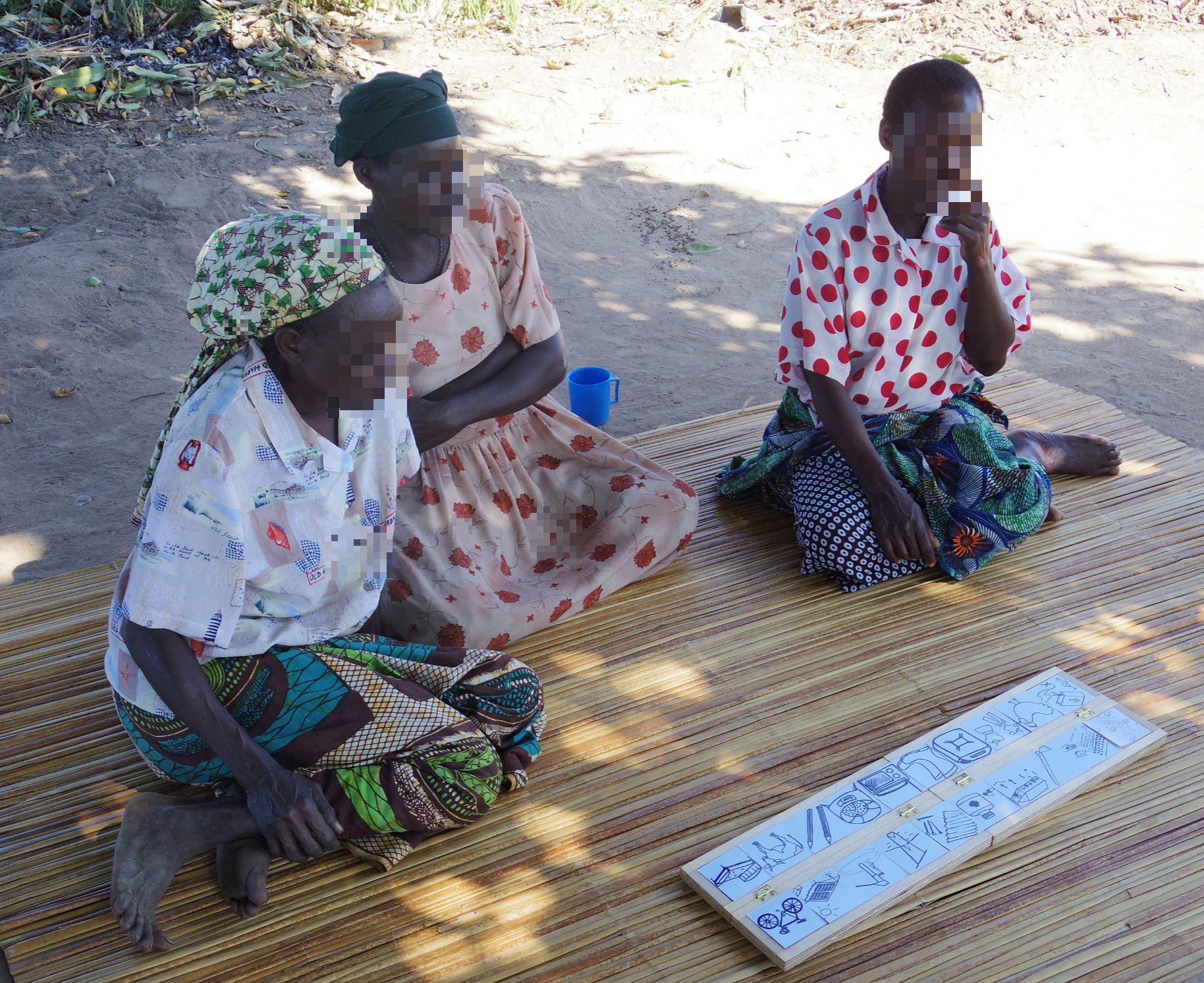}
    \caption{}
    \label{fig:game_women}
 \end{subfigure}
 \hfill
 \begin{subfigure}[b]{0.33\textwidth}
    \centering
    \includegraphics[width=4.6cm]{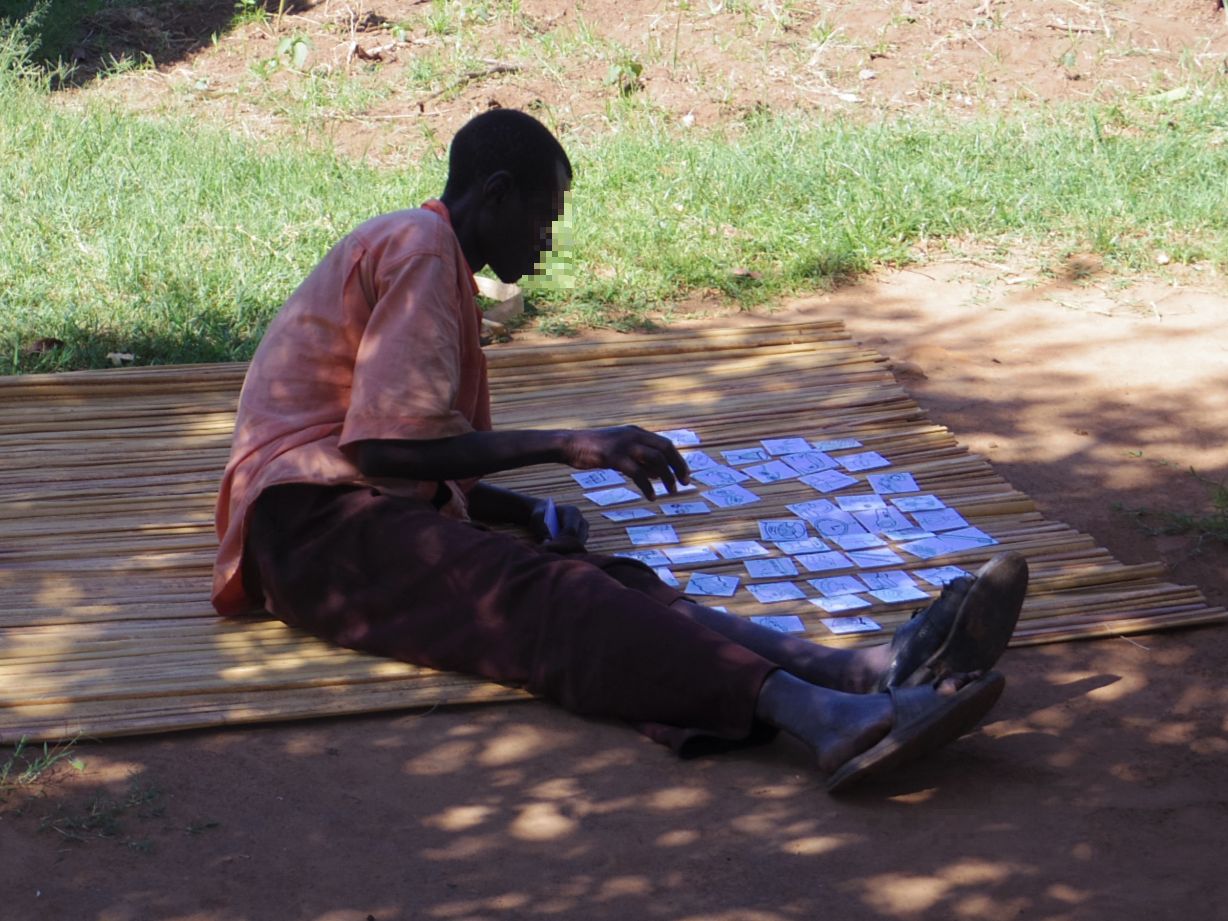}
    \caption{}
    \label{fig:game}
 \end{subfigure}
 \caption{Collecting oral data in rural Uganda.
 \ref{fig:visualprompt}~\textit{Priming effect} (note the word ``Energy'' in the poster's title and the visual prompts differences between items).
 On the contrary,~\ref{fig:game_women} and \ref{fig:game} show minimal priming; note also that different demographics are separately interviewed (women group, single men) to avoid \textit{social desirability bias}.
 }
\end{figure*}

\textbf{Study design}.~
\textcolor{black}{In rural LIC communities}, qualitative data \textcolor{black}{like natural language} is usually collected by observation, interview, and/or focus group discussion (or a combination, known as {mixed methods}) which are transcribed verbatim \cite{moser2018series}. 
Prompts are often used to spark discussion. Whether visual prompts 
\cite{hirmerPhD}
or verbalised question prompts are used \textcolor{black}{during data collection}, these should be designed to: (i)~accommodate illiteracy, (ii)~account for disabilities (e.g.~visually impairment; both could cause \textit{sampling bias}), and (iii)~ minimise bias towards a topic or sector (e.g. minimising \textit{acquisition bias} and \textit{confirmation bias}). For instance, visual prompts should be graphically similar and contain only visuals familiar to the respondents.
\textcolor{black}{This is analogous to the uniform interface with which speakers interact during text-based online data collection, where 
the platform used
is graphically the same to all users inputting data.} Using varied graphical styles or unfamiliar images may result in \textit{priming} (Figure~\ref{fig:visualprompt}). 
To minimise \textit{recall bias} or \textit{recency effect} in collected data, socio-economic data can be integrated in data analysis 
to better understand if the assertions made in collected data reference recent events, for example. These should be non-sector specific, to gain holistic insights and to minimise \textit{acquisition bias} and \textit{confirmation bias}.

\subsection{Engagement}
\label{sec:4_2}

Here, we outline practical steps for successful community engagement to achieve ethical and high-quality data collection. 

\textbf{Defining community}. Defining a community in an open and participatory manner is critical to meaningful engagement ~\cite{Dyer2014assess}. By understanding the community the way they understand themselves, misunderstandings and tensions \textcolor{black}{that affect data quality} can be minimized. 
The definition of the community \cite{MacQueen2001community} coupled \textcolor{black}{with the requirements and use-cases for the collected data}
determines the 
\textcolor{black}{data collection methodology and style which will be most appropriate}
(e.g. interview-based community consultation vs. collaborative co-design for mutual learning).

\textbf{Follow formal structures}. 
Researchers entering a community where they have no background \textcolor{black}{to collect data} should endeavour to know the community prior to commencing any work \cite{Diallo2005community}. This could entail visiting the community and 
mapping 
its hierarchies of authority and decision-making pathways, which can 
guide the research team on how to interact respectfully with the community \cite{Tindana2011align}.
This process should also illuminate whether 
knowledgeable community members should facilitate entry by performing introductions and assisting the external \textcolor{black}{data collection} team. 
Following formal community structures is vital, especially in developing communities, where traditional 
rules and social conventions are strongly held yet often not articulated explicitly or documented. Approaching community leaders in the traditional way 
can help to build a positive long-term relationship,
removing suspicion about the nature and motivation of the researchers' activities, explaining their presence in the community, and most importantly building trust as they are granted permission to engage the community by its leadership~\cite{Tindana2007grand}. 

\textbf{Verbalising consent}.
Data ethics is paramount for research involving human participants~\cite{Accenture2016, Tindana2007grand}, \textcolor{black}{including any collection of personal and identifiable data, such as natural language}.
 Genuine (i.e. voluntary and informed) consent must be obtained from interviewees to prevent use of data which is illegal, coercive, or
 for a purpose other than that which has been agreed
 ~\cite{mcadam2004ethics}.
The Nuffield Council on Bioethics (\citeyear{Bioethics2002}) caution that in 
LICs, misunderstandings may occur due to cultural differences, lower social-economic status, and illiteracy~\cite{McMillan2004ethics} which can call into question the legitimacy of consent obtained.
Researchers must understand that methods 
such as long information forms and consent forms which must be signed may be inappropriate for the cultural context of 
LICs and can be more likely to confuse than to inform
\cite{Tekolaconsent}. 
The authors
advise that consent forms should be verbal instead of written, with wording familiar to the interviewees and appropriate to their level of comprehension \cite{Tekolaconsent}. For example, to speak of data storage on a password protected computer while obtaining consent in a rural community without access to electricity or information technology is unfitting. Innovative ways to record consent can be employed in such contexts (e.g.~video taping or recording), as signing an official document may be ``viewed with suspicion or even outright hostility'' \cite{upjohn2016challenges}, or seen as ``committing ... to something other than answering questions''. Researchers new to qualitative 
data collection should seek advice from experienced researchers and approval from their ethics committee before implementing consent processes. 

\textbf{Approaching participants}. 
Despite having gained permission from community authorities and obtained consent \textcolor{black}{to collect data}, researchers must be cautious when approaching participants \cite{Irabor2009local, Diallo2005community} to ensure they do not violate cultural norms.
For example, in some cultures 
a senior family member must be present for another household member to 
be interviewed, or a female must be accompanied by a male counterpart \textcolor{black}{during data collection}. Insensitivity to such norms may compromise the data collection process; so, they should be carefully noted when researching local context ($\mathsection$\ref{prep}) and interviews should be designed to accommodate them where possible. 
Furthermore, researchers should investigate the motivations of the participants to identify when inducements become inappropriate and may lead to either harm or data bias \cite{mcadam2004ethics}. 

\textbf{Minimise external influence.}~Researchers must be aware of how 
external influences can affect data collection \cite{Jones2012factors}. 
We find three main levels of external influence: (i)~technologies unfamiliar to a rural developing country context may induce \textit{social desirability bias} or \textit{priming} (e.g.~if a researcher arrives to a community in an expensive 
vehicle or uses a tablet for data collection); (ii)~intergroup context, which according to \citet{abrams2010processes} refers to when “people in different social groups view members of other groups” and may feel prejudiced or threatened by 
these differences. This can occur, for instance, when a newcomer arrives and speaks loudly relative to the indigenous community, which may be perceived as overpowering; (iii)~there is the risk of a researcher over-incentivizing the data collection process, using leading questions and judgemental framing (\textit{interviewer bias} or \textit{confirmation bias}).
To overcome these influences, researchers must be cognizant of their influence and minimise it by hiring local mediators where possible alongside employing appropriate technology, mannerisms, and language. 

\footnotetext{While participants' photographing permission was granted, photos were pixelised to protect identity.}

%

\subsection{Undertaking Interviews}

Here, we detail practical steps to minimise challenges during the actual data collection. 

\textbf{Interview settings}.~People have personal values and drivers that may change in specific settings. 
For example, in the Ugandan Buganda and Busoga tribes, it is culturally appropriate for the male head if present to speak on behalf of his wife and children. \textcolor{black}{This could lead to corpora where 
input from the husband is over-represented compared to 
the rest of the family}.
To account for this, it is important to 
\textcolor{black}{collect data}
in multiple \textcolor{black}{interview} settings (e.g.~individual, group male/female/mixed; Figures~\ref{fig:game_women}, \ref{fig:game}).
Additionally, the inputs of individuals \textcolor{black}{in group settings} should be considered independently to ensure all participants have an equal say, regardless of their position within the group~\cite{barry2008determining, Gallagher1993}.
This helps to avoid \textit{social desirability bias} \textcolor{black}{in the data} and is particularly important in various developing contexts where stereotypical gender roles are prominent~
\cite{hirmerPhD}.
%
During interviews, verbal information can be supplemented through the observation of tone, cadence, gestures, and facial expressions
~\cite{Narayanasamy2009, hess2009intergroup}, 
\textcolor{black}{which could enrich the collected data with an additional layer of annotation}.


\textbf{Working with multiple interviewers}.~Arguably, one of the biggest challenges in data collection is ensuring consistency when working with multiple interviewers.
%
Some may report word-for-word what is being said, 
while others may summarise or misreport, resulting in systematic \textit{misunderstanding}.
Despite these risks, employing multiple interviewers is often unavoidable when \textcolor{black}{collecting data} in rural areas of developing countries, where languages often exhibit a high number of regional, non-mutually intelligible varieties.
This is particularly prominent across SSA. For example, 41 languages are spoken in Uganda~\cite{nakayiza2016sociolinguistic}; English, the official language, is fluently spoken by only 
$\sim$5\% of the population, despite being widely used among researchers and NGOs~\cite{katushemererwe2015computer}.
%
%
To minimise data inconsistency, researchers should:
(i)~undertake interviewer training workshops to communicate data requirements and practice data collection processes through mock field interviews;
(ii)~
pilot the data collection
process and seek feedback to spot early deviation from data requirements;
(iii)~regularly spot-check interview notes; 
(iv)~support written notes with audio recordings\footnote{Relying only on audio data recording may be risky: equipment can fail or run out of battery 
(which is not easily remedied in rural off-grid regions) and seasonal factors (as noise from rain on corrugated iron sheets, commonly used for roofing in SSA) can make recordings inaudible~\cite{hirmerPhD}).
};
%
and (v)
\textcolor{black}{
offer quality based incentives to data collectors}.

\textbf{Participant remuneration}.~While it is common to offer interviewees some form of remuneration for their time, 
the decision surrounding payment is ethically-charged and widely contested \cite{Hammett2012}. Rewards may tempt people to participate \textcolor{black}{in data collection} against their 
judgement. They can introduce \textit{sampling bias} 
or create power dynamics resulting in \textit{acquiescence bias} \cite{largent2017paying}. \citet{barbour2013introducing} offers three practical solutions: (i) not advertise payment; (ii) omit the amount being offered; \textcolor{black}{or (iii) offer non-financial incentives (e.g. products that are desirable but difficult to get in an area).}
The decision whether or not to remunerate should not be based upon the researcher's own ethical beliefs and resources, but instead by considering the specific context\footnote{In rural Uganda, for example, politicians commonly engage in \textit{vote buying} by distributing gifts~\cite{blattman2019eat} such as 
soap or alcohol. It is therefore considered an {unruly} form of remuneration and can only be avoided when known.}, interviewee expectations, precedents set by previous researchers, and local norms \cite{Hammett2012}. Representatives from local organisations (such as NGOs or governmental authorities) may be able to offer advice. 

\subsection{Post-interviewing} 

Here, we discuss practical strategies to mitigate ethical issues surrounding the management and stewardship of collected data.
  
\textbf{Anonymisation}.~To protect the participants' identity \textcolor{black}{and data privacy}, locations, proper names, and culturally explicit aspects (such as tribe names) \textcolor{black}{of collected data} should be made anonymous~\cite{sweeney2000simple, kirilova2017rethinking}. This is particularly important in countries with security issues and low levels of democracy. 

\textbf{Safeguarding data}.~A primary responsibility of the researcher is to safeguard participants' data 
\cite{kirilova2017rethinking}.
In addition to anonymizing data, mechanisms for data management include in-place handling and storage of data~\cite{UKRI:data_protection}. 
Whatever data management plan is adopted, it must be clearly articulated to participants before the start of the interview (i.e. as part of the consent process \cite{Silverman2013}), as was discussed in~$\mathsection$\ref{sec:4_2} (\textit{Verbalising consent}). 
%

\textbf{Withdrawing consent}.
Participants should have the ability to withdraw from research within a specified time frame.
This is known as \textit{withdraw consent} and is commonly done by phone or email~\cite{UKRI:withdraw}.
%
As people in rural illiterate communities have limited means and technology access, 
a local phone number and contact details of a responsible person in the area should 
be provided to facilitate withdraw consent.

\textbf{Communication and research fatigue}.
While researchers frequently extract knowledge \textcolor{black}{and data} from communities, only rarely are findings fed back to communities in a way that can be useful to them.
Whatever the research outcomes, researchers should share the results with participating communities in an appropriate manner. In illiterate communities, for instance, murals ~\cite{yale}, artwork, speeches, or song 
could be used to communicate findings.
%
Not communicating findings may result in research fatigue as people in \textit{over-studied} communities are no longer willing to participate \textcolor{black}{in data collection}. This is common ``where repeated engagements do not lead to any experience of change [...]''~\citet{clark2008we}. 
\citet{patel2020research} offers practical guidance 
to minimise research fatigue by: (i) increasing transparency of research purpose at the beginning of the research, and (ii) engaging with gatekeeper or oversight bodies to minimise number of engagements per participant. 
%
Failure to restrict the number of times that people are asked to participate in studies risks poor future participation~\cite{patel2020research} which 
can also lead to \textit{sampling bias}.






\section{Conclusion}\label{conclusion}
In this paper, we provided a first step towards defining best practices in data collection in rural and illiterate communities in Low-Income Countries to create globally representative corpora.
We proposed a comprehensive classification of sources of bias and unethical practices that might arise in the data collection process, and discussed 
practical steps to minimise their negative effects.
%
%
We hope that this work will motivate NLP practitioners to include input from rural illiterate communities in their research, and facilitate smooth and respectful interaction with communities during data collection. 
Importantly, despite the challenges that working in such contexts might bring, the effort to build substantial and 
high-quality corpora which represent this subset of the population can result in
considerable SD outcomes.

\section*{Acknowledgments}
We thank the anonymous reviewers for their constructive feedback.
We are also grateful to Claire McAlpine, as well as Malcolm McCulloch and other members of the Energy and Power Group (University of Oxford) for providing valuable feedback on early versions of this paper. This research was carried out as part of the Oxford Martin Programme on Integrating Renewable Energy. Finally, we are grateful to the Rural Senses team for sharing experiences on data collection.

\bibliography{emnlp2020}
\bibliographystyle{acl_natbib}

\end{document}